\documentclass[runningheads]{llncs}
\usepackage[T1]{fontenc}
\usepackage{graphicx}
\usepackage{amsmath}
\usepackage{amsfonts}
\usepackage[table]{xcolor}
\usepackage{amssymb}
\usepackage{sidecap}
\def\be{\begin{equation}}
\def\ee{\end{equation}}
\def\bea{\begin{eqnarray}}
\def\eea{\end{eqnarray}}

\begin{document}
\title{Evidential Deep Learning with conditional entropy for glaucoma detection}
%
%
\author{Hai Siong Tan \inst{1}\orcidID{0000-0002-6926-1239} \and
Kuancheng Wang \inst{2}
}
\authorrunning{H.S.Tan and K.Wang}
%
\institute{Gryphon Center for Artificial Intelligence and Theoretical Sciences, Singapore \email{haisiong.tan@gryphonai.com.sg} 
\and
Electrical and Computer Engineering Department, \\University of California San Diego, CA, USA
}
\maketitle              
\begin{abstract}
We propose a novel conditional entropy function as a basis for quantifying predictive uncertainty in Evidential Deep Learning (EDL) for glaucoma detection from retinal fundus images. Using the EyePACS-AIROGS-light-V2 and RIM-ONE-DL datasets for validation, we found that our proposed EDL model regularized by conditional entropy function produced uncertainty estimates with substantially lower
calibration errors than Deep Ensembles, MC Dropout, post-hoc calibration and the conventional EDL framework based on Dempster uncertainty. To further probe our models' interpretability, we analyzed Grad-CAM heatmaps using a focus index representing spatial concentration. Compared with the other uncertainty quantification methods, EDL models exhibited greater separation between the index distributions of correctly and incorrectly classified fundus images. In particular, false negatives were associated with localized Grad-CAM activations that were displaced from the optic disc, consistent with its established clinical importance in glaucoma assessment. Overall, our findings suggest that our proposed EDL framework for glaucoma detection provides not only enhanced uncertainty calibration but also admits a more physiologically consistent interpretability.

\end{abstract}
\section{Introduction}
\label{sec:intro}

This manuscript is devoted to the subject of using neural network models to recognize glaucoma -- a group of ocular conditions characterized by progressive damage to the optic nerve -- from retinal fundus images~\cite{shields}. The effectiveness of using deep learning techniques for glaucoma detection has been explored primarily through the use of transfer learning~\cite{Camara,Dulani}, where established models trained on large database such as ImageNet are used and their weights adjusted to the smaller fundus image datasets during model training. Past works in these areas indicated that using neural network models for these purposes is effective as a complementary method for glaucoma diagnosis, and more generally other eye diseases such as age-related macular degeneration~\cite{mD} and diabetic retinopathy~\cite{dR}. 

A major theme of our work here lies in uncertainty quantification (UQ). To our knowledge, the majority of literature on neural network-assisted glaucoma detection has been focused on the comparative performances of various models defined with regards to accuracy-related metrics such as AUROC, etc.~\cite{Camara,Dulani}, but lacking analysis of the reliability of model predictions. In a safety-critical application like glaucoma screening, it would be ideal for model's predictions to be accompanied by some measure of their uncertainties or level of model confidence, so that users are aware of some measure of the statistical reliability behind each image classification. For any model prediction associated with a relatively large uncertainty, physicians would be motivated to trust it to a lesser extent and rely more on their own manual assessment. Having a sense of how confident the model is in predicting glaucoma from the fundus image thus enables a healthier reliance on the model. Ideally, the model-generated probability of the image indicating glaucoma should also be consistent empirically with its mean accuracy, so that we can trust the probabilistic output and potentially leverage it for tracking rate of disease progression. Thus, in this work, our primary metric is the expected calibration error~\cite{jungo} which has been an essential measure of UQ in various areas of machine learning~\cite{uq_review}. 

Within the landscape of UQ methods, our study compares among four main frameworks: Evidential Deep Learning (EDL)~\cite{sensoy}, Monte Carlo Dropout~\cite{gal}, post-hoc temperature scaling and Deep Ensemble~\cite{lak}, leaving a more extensive investigation for future work. For EDL, we propose a novel conditional entropy function that can be used to quantify predictive uncertainty instead of Dempster uncertainty in standard EDL formalism. This is essentially derived in the spirit of Bayesian statistics by marginalizing the Shannon entropy with respect to the prior density function. We found that when EDL model training was regularized by our entropy function, it led to superior performance compared to the standard formalism and other UQ frameworks.  Theoretically, it amounts to a re-interpretation of EDL as an UQ technique that can be solely understood using conventional statistical and Bayesian principles without alluding to Dempster-Shafer theory of evidence~\cite{DS,shafer}.
Our paper is organized as follows. 
In Sec.~\ref{sec:method}, we describe our novel entropy function, its associated regularizer term and various aspects of model implementation. Sec.~\ref{sec:results} presents the main results of our experiments which are summarized in Table~\ref{tab:origa_results}. We discuss clinical implications of our results in Sec.~\ref{sec:discussion} before concluding in Sec.~\ref{sec:conclusion}. The Appendix provides a brief analysis of the asymptotic behavior of the entropy function. We hope that our work here contributes towards robust clinical deployment of A.I.-enabled glaucoma detection from fundus images.

\section{Methodology}
\label{sec:method}

\subsection{On datasets}
The two retinal image datasets that we have chosen for model training for this work are representative of two major classes of fundus images - (i)full fundus that presents the complete, wide-angle picture of the retina, macula and blood vessels, and (ii)cropped fundus that is zoomed onto the neighborhood of the optic nerve head, including the optic disc region.

\begin{enumerate}
\item \textbf{EyePACS-AIROGS-light-v2} set: 
This is a machine-learning-ready full-fundus glaucoma dataset curated in \cite{eyepacs2,E2} using a balanced subset of standardized fundus images from the Rotterdam EyePACS AIROGS set \cite{airogs}. Image sampling and standardized preprocessing techniques were applied to the original Rotterdam dataset for more convenient benchmarking with reduced dataset size. This dataset is split into training, validation, and test folders which contain 4000 (84\%), 385 (8\%), and 385 (8\%) fundus images in each class respectively. 

\item The \textbf{RIM-ONE-DL} image dataset \cite{RIM_ONE} consists of 313 fundus images from normal subjects and 172 images from patients with glaucoma. These cropped fundus pictures were collected in three Spanish hospitals and cropped squarely around the head of the optic nerve using the same proportionality criterion. For benchmarking purposes, the authors of \cite{FUM} have performed a specific train/test split that contain 339 (70$\%$) and 146 (30\%) fundus images. For the validation dataset used in our hyperparameter ablation experiments, we further split the testing dataset into two.
\end{enumerate}

Apart from these two main datasets, we also leveraged two well-known retinal fundus dataset: (i)the \textbf{CRFO} dataset of \cite{hassan} and (ii)the \textbf{ACRIMA} dataset of \cite{acrima} to assess the effectiveness of using uncertainty estimates for identifying out-of-distribution (OOD) images. The CRFO dataset contains 79 full-fundus images (containing 48 glaucoma cases) while the ACRIMA dataset contains 705 cropped images (containing 396 glaucoma cases). These datasets exhibit systematic distributional shifts in the LAB luminance channel relative to EyePACS and RIM-ONE, respectively. We leveraged these distributional shifts to assess OOD detection performance using the AUROC computed from the uncertainty estimates of different models.

\subsection{On Evidential Deep Learning}

A fundamental idea behind Evidential Deep Learning (EDL) as proposed in the seminal work~\cite{sensoy} is the specification of a prior distribution for the discrete probability distribution assumed to underlie the classification task. For a binary class problem, this is the Bernoulli distribution and in EDL, it is assumed to have Beta distribution as its prior, of which density function reads
\be
\label{beta_prior}
\pi \left(p_1, p_2 | \alpha, \beta\right) = 
\frac{p^{\alpha - 1}_1 p_2^{\beta - 1}}{B(\alpha, \beta)}, \,\,\,
B(\alpha, \beta) \equiv \frac{\Gamma(\alpha)\Gamma(\beta)}{\Gamma(\alpha + \beta)},
\ee
where $p_i$ denotes the class probability with the constraint $p_1 +p_2 = 1$, $\Gamma (\alpha)$ denotes the Gamma function and $\alpha, \beta \geq 1$ are prior parameters that are used to express various quantities of interest. For example, the mean probability is obtained after integrating over the prior density
\be
\langle p_j \rangle = 
\int^1_0 \, dp_j 
\, p_j \frac{p^{\alpha - 1}_1 p_2^{\beta - 1}}{B(\alpha, \beta)}
= \frac{\alpha_j}{\alpha + \beta}, \,\,\, j = 1,2.
\ee
A similar approach is used for deriving a suitable loss function, for example integrating the standard cross-entropy loss or the mean-squared-error loss over the Beta prior yields eqns. 4 and 5 of \cite{sensoy}. As for the predictive uncertainty, authors of the original EDL framework \cite{sensoy} alluded to Dempster-Shafer theory~\cite{DS,shafer} to \emph{assert} that the quantity $u_d = 2/(\alpha + \beta)$ represents uncertainty -- in the sense that we \emph{interpret} each of the two bracketed terms in the following equality as measures of the amount of data support (or `belief mass' in the language used in \cite{sensoy}) in favor of the classification. 
\be
\label{dempster_l}
u_d + \left( \frac{\alpha -1 }{\alpha + \beta} \right) + \left( \frac{\beta - 1}{\alpha + \beta} \right) = 1.
\ee
Although Dempster-Shafer Theory~\cite{DS,shafer} was cited for the interpretational postulate underlying eqn.~\eqref{dempster_l}, apart from semantics and literal definitions, we are unaware of any conventional information-theoretic justification for taking $u_d$ as a pure measure of uncertainty. In this work, we allude to an entropy-based definition of uncertainty instead. 
Similar to how the mean probabilities and loss functions formulated in \cite{sensoy} were derived from integrating functions of the class probabilities over their Beta priors, a natural notion of uncertainty can be derived in the Bayesian sense by marginalizing the uncertainty measure with respect to the prior density. 

\subsection{Deriving a coarse-grained entropy for EDL models}
 
One common measure of uncertainty is the Shannon entropy. Denoting the probability of the $k^{\text{th}}$ image being classified as glaucomatous by $p_k$, the Shannon entropy representing probabilistic uncertainty associated with the binary classification is the binary entropy function defined as 
\be
\label{shannon}
S = -\sum_k \left( p_k \log\,p_k + (1 - p_k)
\log \, (1-p_k) \right).
\ee
In the original EDL framework~\cite{sensoy}, the loss function was derived by averaging the mean-squared-error term over $p$ taking the prior for $p$ to be the Beta distribution parametrized by $\alpha, \beta$. In a similar spirit, we consider the average binary entropy $S_c$ obtained by marginalizing \eqref{shannon} with respect to the prior density. 
\be
\label{f1}
S_c = - \sum_k \int^1_0 
dp_k
\,  \Bigg( p_k \log\,p_k + (1 - p_k)
\log \, (1-p_k) \Bigg)
\frac{p^{\alpha - 1}_k (1-p_k)^{\beta - 1}}{B(\alpha, \beta)}, 
\ee
where $B(\alpha, \beta) \equiv
\frac{\Gamma(\alpha)\Gamma(\beta)}{\Gamma(\alpha + \beta)}$ is the Beta function. Using the integration formula~\cite{formula}
$$
\int^1_0 dx\, \log(x)\,
x^a (1-x)^b = 
\frac{\Gamma(1+a)\Gamma(1+b)( H_a - H_{1+a+b} )}{\Gamma(2+a+b)},\,\,\,
a>-1,b>-1, 
$$
where $H_a$ denotes harmonic numbers analytically continued to the real plane, we obtain
\be
\label{sm}
S_c = \sum_k \psi(\alpha_k + \beta_k + 1) - \frac{\alpha_k \psi(\alpha_k + 1) + \beta_k \psi(\beta_k + 1)}{\alpha_k + \beta_k},\,\,\, \psi(x) \equiv \Gamma'(x)/\Gamma(x), 
\ee
where $\psi$ is the digamma function.\footnote{The digamma functions are related to
the analytically continued harmonic numbers via 
$H_\alpha = \psi(\alpha + 1) + \gamma$, where $\gamma$ is the Euler-Mascheroni constant.} Our coarse-grained entropy function admits a natural interpretation in classical information theory --  that of \emph{conditional entropy} which is fundamentally defined~\cite{thomas} as the expected value of entropies of distributions, averaged over the conditioning random variable. In our binary classification context, the parameter $p$ of the Bernoulli distribution is lifted to be a continuous, bounded variable with the Beta distribution as the prior density. Taking this to be the conditioning random variable, the form of eqn.~\eqref{f1} can be identified with that of conditional entropy $H(Y|\mathcal{P})$ with $Y$ being the discrete binary-valued random variable with a Bernoulli distribution conditioned on a continuous, bounded variable $\mathcal{P}$ equipped with a Beta distribution. Each term of the sum in \eqref{f1} can be written as 
\be
H(Y|\mathcal{P}) = \int^1_0 dp \, \pi(p) H(Y| \mathcal{P}=p ) = 
\int^1_0  dp \, \pi(p) \left[
p \log \, p + (1-p) \log \, (1-p)
\right],
\ee
where $\pi(p) = \frac{p^{\alpha - 1} (1-p)^{\beta - 1}}{B(\alpha, \beta)}$ is the Beta distribution. Henceforth, we call $S_c$ the conditional entropy. By inspection, it is positive-definite since the binary entropy function $H(Y|p)$ is non-negative and $S_c$ is defined by integrating $H(Y|p)$ with respect to another normalizable, non-negative density function $\pi(p)$.

\begin{figure}[htbp]
    \centering
    \begin{minipage}[c]{0.6\textwidth}      \includegraphics[width=\textwidth]{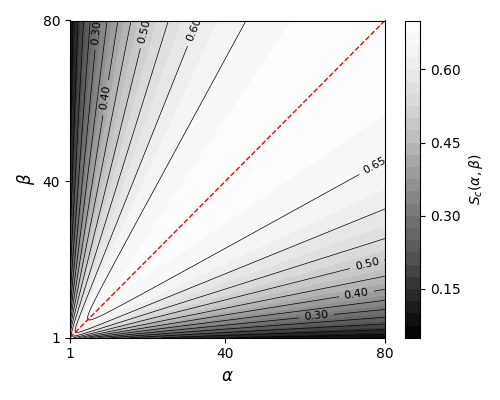}
    \end{minipage}
     \hspace{0.02\textwidth}
    \begin{minipage}[c]{0.28\textwidth}
        \caption{Contour plot of the conditional entropy function $S_c$, with darker regions representing lower values. The red dashed line is $\alpha = \beta$, along which the global supremum of $\log 2$ is approached at infinity. In the Appendix, we describe the asymptotic region in more detail.}
  \label{fig:entropy_plot}
    \end{minipage}
\end{figure}

Following \cite{sensoy}, we take $\alpha,\beta \geq 1$ and in this 
domain, the conditional entropy $S_c$ satisfies 
\be
\label{entropy_bound}
0 < S_c (\alpha,\beta) < \log 2, \,\,\,
\lim_{\substack{\alpha=\beta \\ \alpha \to\infty}} S_c (\alpha,\beta)=\log 2.
\ee
This supremum is approached along $\alpha = \beta \rightarrow \infty$. To interpret this maximal value, we note that the binary entropy function is maximized by $p = \frac{1}{2}$. The limit $\alpha = \beta \rightarrow \infty$ corresponds to the Beta prior collapsing to a singular $\delta$-function at $p = \frac{1}{2}$, yielding the maximum value of the average binary entropy $S_c$. Away from this singular limit, other finite values of $(\alpha,\beta)$ yield lower average entropies since they characterize a probabilistic spread of $p$ instead of just taking it to be $\frac{1}{2}$. 
Fig.~\ref{fig:entropy_plot} is a contour plot of $S_c$ in the domain $(\alpha, \beta) \in [1,\infty)\times [1,\infty)$. 
In the Appendix, we describe the asymptotic region in more detail. 

For $\alpha = \beta = 1$, $S_c = \frac{1}{2}$ -- the value of the average binary entropy when the Beta prior for $p$ is equivalently the uniform distribution $\mathcal{U}(0,1)$ for these values of $\alpha, \beta$. The infimum of $0$ is attained along $\beta=1, \alpha \rightarrow \infty$ or $\alpha = 1, \beta \rightarrow \infty$ which pertain to the Beta prior reduced to singular $\delta$-function localized at $p=1$ or 0 respectively. In summary, the conditional entropy $S_c$ generalizes the binary entropy function by computing the average entropy of the stochastic distribution of $p$ as defined by the Beta prior.

\subsection{A conditional entropy-based regularizer for model training}

To integrate conditional entropy as a measure of predictive uncertainty in model training, a common approach is to regularize model training such that the uncertainty associated with the outputs are consistent with eqn.~\eqref{sm}. Similar in spirit to the original seminal works of EDL~\cite{amini,sensoy}, we use a loss term that increases our measure of uncertainty for erroneous classifications. Since we would like the conditional entropy to be larger for erroneous predictions, we consider the regularizer
\be
\label{L_m}
\mathcal{L}_{c} = \delta^2
\left( 1 - \frac{1}{\log 2}S_c (\alpha, \beta) \right), \,\,\,
\delta \equiv y_t (1 - y_p) + (1-y_t)y_p.
\ee
where $\delta$ represents a differentiable error term with $y_t, y_p$ being ground truth and predicted classes respectively. We choose the loss term to be quadratic in $\delta$ so that it is more sensitive to the wrongly predicted samples. In eqn.~\eqref{L_m}, the $\log 2$ factor normalizes the entropy term so that it is bounded within (0,1). Critically, using the regularizer \eqref{L_m} implements our use of conditional entropy as predictive uncertainty, since $S_c(\alpha, \beta)$ is now invoked directly in model training.

\subsection{On the implementation of EDL models and other UQ frameworks}

MobileNetV3-Small~\cite{MobileNetV3} from the Torchvision library~\cite{PyTorch} was used as the backbone model. The network was initialized with ImageNet-pretrained weights for transfer learning. 
As an adjustment module for our 
glaucoma classification task, the classifier head was replaced with a fully connected multi-layer perceptron (MLP) producing a single output logit which was subsequently passed through a sigmoid activation function to generate probabilities in the baseline model. The MLP was picked to be 3-layer with the number of neurons being $(256, 128, 1)$ for the EyePACS dataset and $(64, 32, 1)$ for the RIM-ONE dataset and model was trained using the binary cross entropy loss. 
For EDL models, a slightly different classification head was used, consisting of a 4-layer MLP with the number of neurons being $(256, 128, 64, 2)$ for the EyePACS dataset and $(128, 64, 32, 2)$ for the RIM-ONE dataset. The final two-unit output layer were taken to be logits which were then passed through a softplus function to generate the evidence vectors $e_j$ defined as $\alpha_j = 1 + e_j$. The loss function in our experiments can be generally expressed as
\be
\mathcal{L}_{EDL} = \lambda_{ce} \mathcal{L}_{ce} + \lambda_{c} \mathcal{L}_{c} + \lambda_{kl}
\mathcal{L}_{kl}, 
\ee
where $\mathcal{L}_{c}$ is defined in eqn.~\eqref{L_m}, and $ \mathcal{L}_{ce},\mathcal{L}_{kl}$ are the cross-entropy and KL-divergence regularizer of \cite{sensoy}\footnote{Detailed formulas for $ \mathcal{L}_{ce},\mathcal{L}_{kl}$ can be found in Section 4 of \cite{sensoy}.}. Models in the standard EDL model are defined as setting $\mathcal{L}_c=0$, whereas our proposed model is associated with setting $\mathcal{L}_{KL} =0$ since this regularizer is defined with respect to Dempster uncertainty instead. 
The coefficients $(\{ \lambda_{ce}, \lambda_{c}, \lambda_{kl}\})$, like the number of hidden neurons per layer of the MLP adjustment heads, were picked after performing ablation experiments. 
After a grid search involving four orders of magnitude (0.01 to 100) for each of $(\lambda_{CE}, \lambda_c, \lambda_{kl})$, the optimal values were $\lambda_{ce}=\{1.0,0.1\},\lambda_c=\{0.15,0.1\},\lambda_{kl}=\{1.0,0.1\}$ for EyePACS-and RIM-ONE-trained models respectively. All models were trained with Adam optimizer and a learning rate of $10^{-4}$ for 50 epochs at which a relative tolerance $\lesssim 0.01$ was attained. 

In this work, we compared the EDL methodology against other standard UQ frameworks -- post-hoc temperature scaling~\cite{temperature}, Deep Ensemble~\cite{lak} and MC Dropout~\cite{gal}. Each of the 10 neural networks in the ensemble shared the same MobileNet architecture (equipped with the MLP adjustment module for classifier head described earlier) with random weight initialization performed via the \emph{He}-uniform initializer~\cite{Heming} to induce different initial weights. For MC Dropout model, we added a dropout layer after each hidden layer of the MLP head attached to the baseline model, and 20 forward passes were used for the Monte Carlo simulation. For post-hoc temperature scaling of the baseline models, the output logits were first cached on the validation data, after which a scalar temperature was optimized by minimizing the Brier score of the resulting sigmoid probabilities. The optimized temperature\footnote{The post-hoc calibrated temperature parameter turned out to be $T=1.71, 1.19$ for the EyePACS and RIM-ONE-trained baseline models respectively.} was used to rescale the logits prior to computing the calibrated probabilities.
Trained on an A100 GPU, we found that generally, EyePACS-based EDL models required 50 minutes for training which was about twice the duration required for RIM-ONE-based models.

\subsection{Some comments on evaluation metrics}
The expected calibration error (ECE) is a measure of how well the model's probabilistic predictions match empirical ones. Splitting the dataset into a number of bins, it is defined as a weighted mean of the absolute difference between the bin-averaged accuracy
and confidence~\cite{jungo}, i.e.
$$
ECE \equiv 
\sum_{\mu = 1}^M
\frac{|B_\mu |}{N}
\left|f_{acc}(B_\mu) - f_{pred}(B_\mu) \right|,
$$
where $N$ is the number of datapoints, $B_\mu$ are the $M$ bins, $f_{acc}, f_{pred}$ are the average accuracies and predicted probabilities in each bin. 
For classification tasks, these quantities are
$
f_{acc} = \frac{1}{|B_\mu|}
\sum_{i\in B_{\mu}} \delta_{\hat{y}_i, y_i},\,
f_{pred} = \frac{1}{|B_\mu|}
\sum_{i\in B_{\mu}} c_i,
$
where $y_i, \hat{y}_i$ are the predicted and ground-truth class labels respectively, with $c_i$ being the probabilistic confidence corresponding to the predicted class, i.e. $c_i = \arg\max_{k} p^k_i$ where $p^k_i$ are the softmax class probabilities for category $k$ predicted for the $i^{\text{th}}$ datapoint. We evaluated ECE using both equal-width ($\equiv \text{ECE}_w $) and equal-quantile binning ($\equiv \text{ECE}_q $) schemes with 10 bins.  

To investigate the spatial regions that influenced the model's predictions, we employed Gradient-weighted Class Activation Mapping (Grad-CAM) \cite{GradCAM} to generate class-discriminative activation maps. To describe the concentration of these activations, we computed a focus index, defined as the normalized complement of the effective number of activated pixels, with larger values indicating more localized attention. 
We first normalize the Grad-CAM values, converting them to an empirical probability distribution $g_i$ over all pixels of each fundus image. This is used to define an effective number of activated pixels (see e.g.~\cite{Anderson})
$
N_{eff} \equiv 1/\sum_i g^2_i,
$
which is commonly used to quantify localization and sometimes known as the inverse participation ratio~\cite{Anderson}. To obtain an intuitive measure of spatial concentration, we define an index
$
F \equiv 1 - \frac{N_{eff}}{N_t},
$
where $N_t$ is the total number of pixels. Larger values of $F$ indicate more spatially concentrated Grad-CAM activations, while smaller values indicate a more uniformly spread distribution.

\section{Results}
\label{sec:results}
\begin{figure}[h]
    \centering  \includegraphics[width=\textwidth]{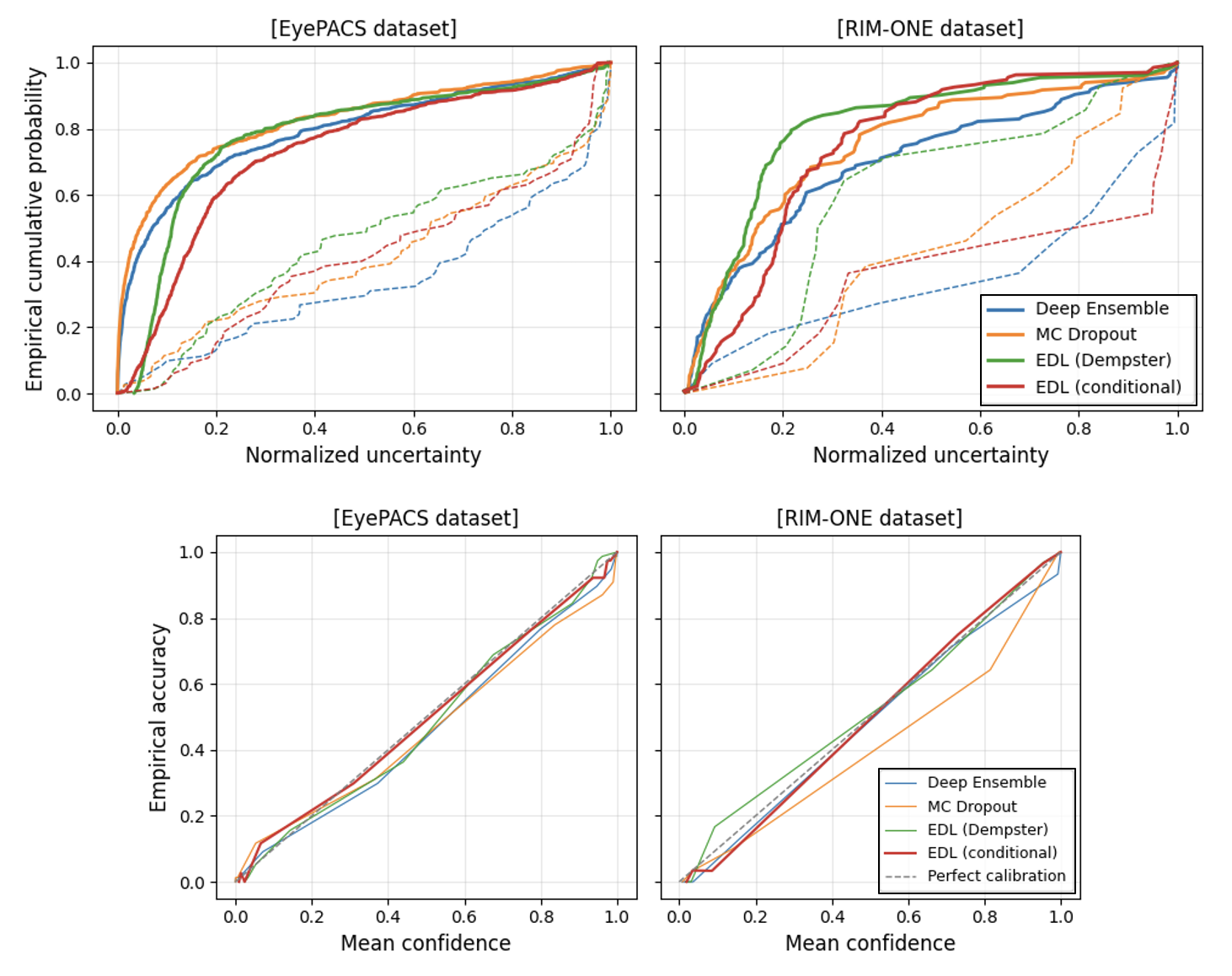}
    \caption{\emph{top}: ECDF of uncertainty estimates for correct (solid) and wrong (dashed) classifications; \emph{bottom}: Reliability diagrams using equal-quantile binning (10 bins). }
\label{fig:ece_plot}
\end{figure}

In Table~\ref{tab:origa_results}, we collect various metrics characterizing model performances. In the aspect of AUC, various models achieved an average value of 0.96 (EyePACS) and 0.97 (RIM-ONE). For accuracy, all models attained values $\gtrsim 0.90$. The EDL models trained with $\mathcal{L}_{c}$ attained the highest accuracy for both datasets, and highest AUC for the RIM-ONE dataset. These differences among them were rather minor ($\lesssim 2\%$) and we regard various models as achieving similar accuracy performance for both datasets. 
While the models exhibited similar accuracy, the ECE metrics revealed more pronounced differences among them, with the baseline model, temperature scaling and MC Dropout being clearly inferior to other models for both datasets. Reliability diagrams for both datasets are shown in Fig.~\ref{fig:ece_plot}. EDL models trained with $\mathcal{L}_c$ attained the lowest ECE for both datasets, and the lowest Brier scores. Deep Ensemble method attained a similar degree of calibration for the RIM-ONE dataset and from Fig.~\ref{fig:ece_plot}, it was relatively overconfident compared to EDL models, whereas MC Dropout models appeared to be the most overconfident ones. In the aspect of uncertainty-based OOD detection, we found the model ranking by this metric to be dataset-dependent. Nevertheless, our proposed EDL model achieved the highest OOD-AUROC for EyePACS and, together with Deep Ensemble, for RIM-ONE. 

\begin{table}[ht]
\centering
\small
\caption{Abbrev -- Acc: accuracy; AUC: area under ROC curve; ECE${}_q$, ECE${}_w$: mean ECE metrics that are quantile-based and fixed width-based; $B_s$: Brier score; $W_h, W_g$ - Wasserstein distance between Grad-CAM indices of correct and wrong predictions for healthy and glaucomatous eye images; OOD: the AUROC score for identifying OOD images based on uncertainty; EDL(Dempster): EDL models trained with $\mathcal{L}_{kl}$ of \cite{sensoy}; EDL(w/o reg.): models trained with $\mathcal{L}_c=\mathcal{L}_{kl}=0$ in the loss function.}
\begin{tabular}{|p{3cm}|c|c|c|c|c|c|c|c|}
\hline
\multicolumn{9}{c}{\textbf{EyePACS dataset-trained model performance}} \\
\hline
\textbf{Models} & Acc & AUC & ECE${}_q$ & ECE${}_w$ & $B_s$ & $W_g$ & $W_h$ & OOD \\ 
\hline
(Baseline) MNetV3 & 0.90 & 0.96 & 0.054 & 0.063 & 0.08 & 0.038 & 0.050 & 0.82 \\ 
\hline
Temp. Scaling & 0.90 & 0.96 & 0.047 & 0.050 & 0.08 & 0.050 & 0.038 & 0.82 \\ 
\hline
MC Dropout & 0.90 & 0.96 & 0.038 & 0.044 & 0.08 & 0.103 & 0.043 & 0.79\\ 
\hline
Deep Ensemble & 0.91 & \cellcolor{gray!20}0.97 & 0.025 & 0.030 &\cellcolor{gray!20}0.07 & 0.105 & 0.138 & 0.81\\ 
\hline
EDL (w/o reg.) & 0.90 & 0.96 & 0.029 & 0.033 & 0.08 & 0.42 & 0.49 & 0.73\\ 
\hline
EDL (conditional) &\cellcolor{gray!20}0.92 & 0.96 &\cellcolor{gray!20}0.019 &\cellcolor{gray!20}0.019 & \cellcolor{gray!20}0.07 & 0.28 & 0.40 
& \cellcolor{gray!20}0.87\\ 
\hline
EDL (Dempster) & 0.89 & 0.95 & 0.025 & 0.026 & 0.09 & \cellcolor{gray!20}0.51 & \cellcolor{gray!20}0.56 & 0.75 \\ 
\hline
\hline
\multicolumn{9}{c}{\textbf{RIM-ONE dataset-trained model performance}} \\
\hline
\hline
\textbf{Models} & Acc & AUC & ECE${}_q$ & ECE${}_w$ & $B_s$ & $W_g$ & $W_h$ & OOD\\ 
\hline
(Baseline) MNetV3 & 0.91 & 0.97 & 0.048 & 0.062 & 0.07 & 0.55 & 0.56 & 0.60 \\ 
\hline
Temp. Scaling & 0.91 & 0.97 & 0.056 & 0.056 & 0.07 & 0.56 & 0.55 & 0.66\\ 
\hline
MC Dropout & 0.91 & 0.97 & 0.045 & 0.057 & 0.08 & 0.40 & 0.56 & 0.68\\ 
\hline
Deep Ensemble & \cellcolor{gray!20}0.92 & 0.97 & 0.027 & 0.035 & 0.06 & 0.49 & 0.56 & \cellcolor{gray!20}0.70\\ 
\hline
EDL (w/o reg.)&\cellcolor{gray!20}0.92 & 0.97 & 0.035 & 0.040 &\cellcolor{gray!20}0.05 & 0.77 &\cellcolor{gray!20}0.66  & 0.68\\ 
\hline
EDL (conditional) & \cellcolor{gray!20}0.92 & \cellcolor{gray!20}0.98 & \cellcolor{gray!20}0.021 & \cellcolor{gray!20}0.033 & \cellcolor{gray!20}0.05 & 0.79 & 0.63 & \cellcolor{gray!20}0.70\\ 
\hline
EDL (Dempster) & 0.90 & 0.96 & 0.032 & 0.036 & 0.08 & \cellcolor{gray!20}0.80 & 0.61 & 0.69\\ 
\hline
\end{tabular}
\label{tab:origa_results}
\end{table}

 While small values of the ECE metrics formally quantify how robust the probabilistic outputs are, another useful feature of the uncertainty distribution is that it may provide some degree of discrimination between correct and wrong predictions. In Fig.~\ref{fig:ece_plot}, we plot the empirical cumulative distribution functions (ECDF) versus the uncertainty estimates for various models trained on each dataset. They depict how uncertainty distributions for the sets of wrong predictions were all markedly separate from those for the correct sets, with the separation stronger for experiments involving the larger EyePACS dataset.

\begin{figure}[ht]
    \centering  \includegraphics[width=\textwidth]{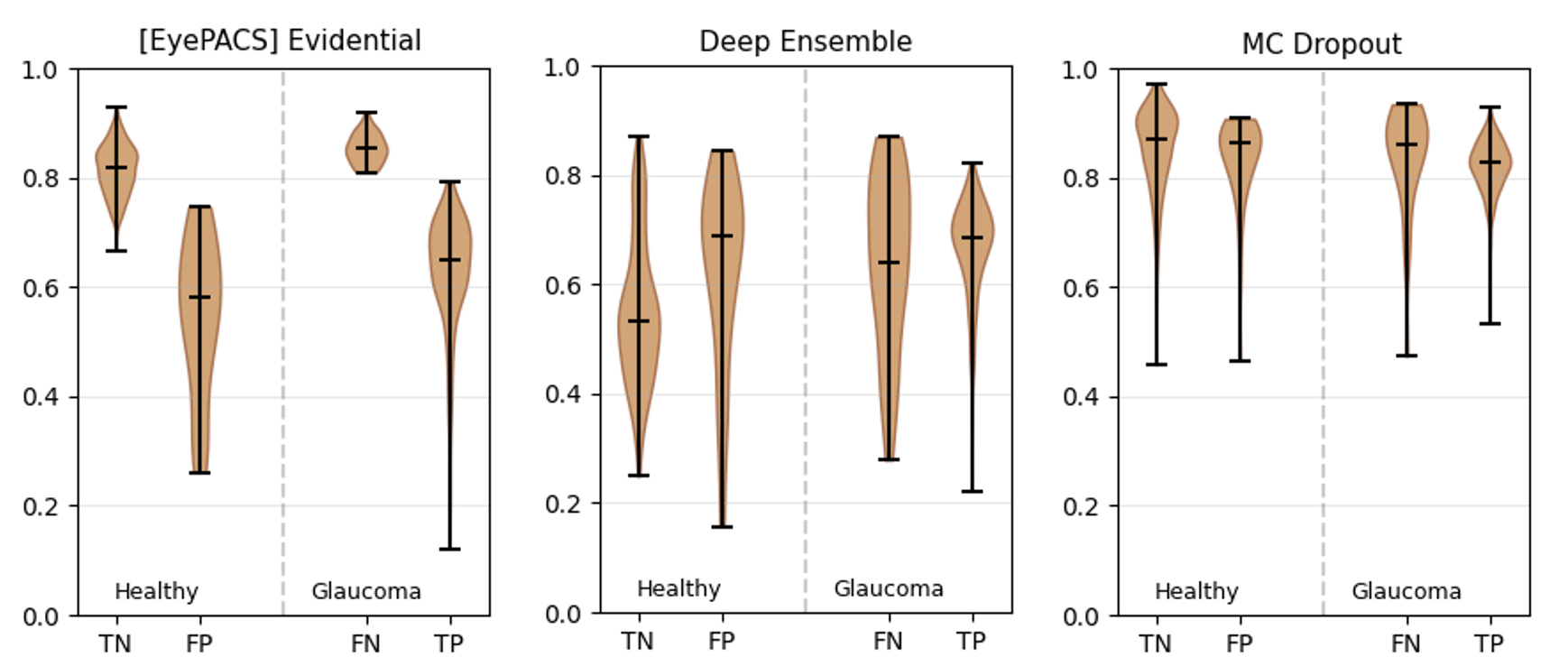}
    \caption{Violin plots of the focus index characterizing Grad-CAM heatmaps of EDL, Deep Ensemble, and MC Dropout models trained on the EyePACS dataset. The EDL model here was trained with $\mathcal{L}_c$, although the distinct separation between correct and wrong predictions also characterized other EDL models (see Wasserstein distances of Table~\ref{tab:origa_results}). Similar relative trends were also observed for RIM-ONE-trained models.}
\label{fig:eyepacs_violin}
\end{figure}

Beyond predictive accuracy and uncertainty-based metrics, we examined the Grad-CAM focus indices to characterize the spatial behavior of various models. Evidential models consistently exhibited greater separation between correct and incorrect predictions in terms of the focus index distributions than the other UQ approaches, as reflected by the Wasserstein distances between the corresponding distributions tabulated in Table~\ref{tab:origa_results}. Fig.~\ref{fig:eyepacs_violin} shows the violin plots of focus indices for three classes of models. While Deep Ensemble and MC Dropout produced substantially overlapping focus index distributions -- a trend also displayed by the baseline and post-hoc temperature scaling, the EDL models demonstrated clearly more pronounced index separations between correct and incorrect predictions. This feature was found to characterize general EDL models, regardless of whether they were trained with $\mathcal{L}_c$ or $\mathcal{L}_{kl}$.


\section{Discussion}
\label{sec:discussion}

The EDL models regularized by conditional entropy appeared to demonstrate consistent superiority in uncertainty calibration (as measured by ECE${}_q$ and ECE${}_w$). Other models were weaker in their degrees of uncertainty calibration being mostly overconfident as apparent in Fig.~\ref{fig:ece_plot}. In the absence of any regularizer, the EDL models did not exhibit advantages over standard ones trained using a regularizer $\mathcal{L}_{kl}$ of \cite{sensoy} that is based on Dempster uncertainty. The ECDF plots in Fig.~\ref{fig:ece_plot} show
the uncertainty-aware models to display various degrees of separation in the uncertainty distributions between correct and wrong predictions. In the aspect of Grad-CAM indices, we found that EDL models exhibited a property quite distinct from Deep Ensemble, temperature scaling and Monte Carlo Dropout -- the focus index distributions pertaining to correct and wrong predictions were clearly separable for both datasets. In particular, within the set of glaucomatous images, the indices for false negatives spread around higher medians relative to true positive images. 

\begin{figure}[h!]
    \centering  \includegraphics[width=\textwidth]{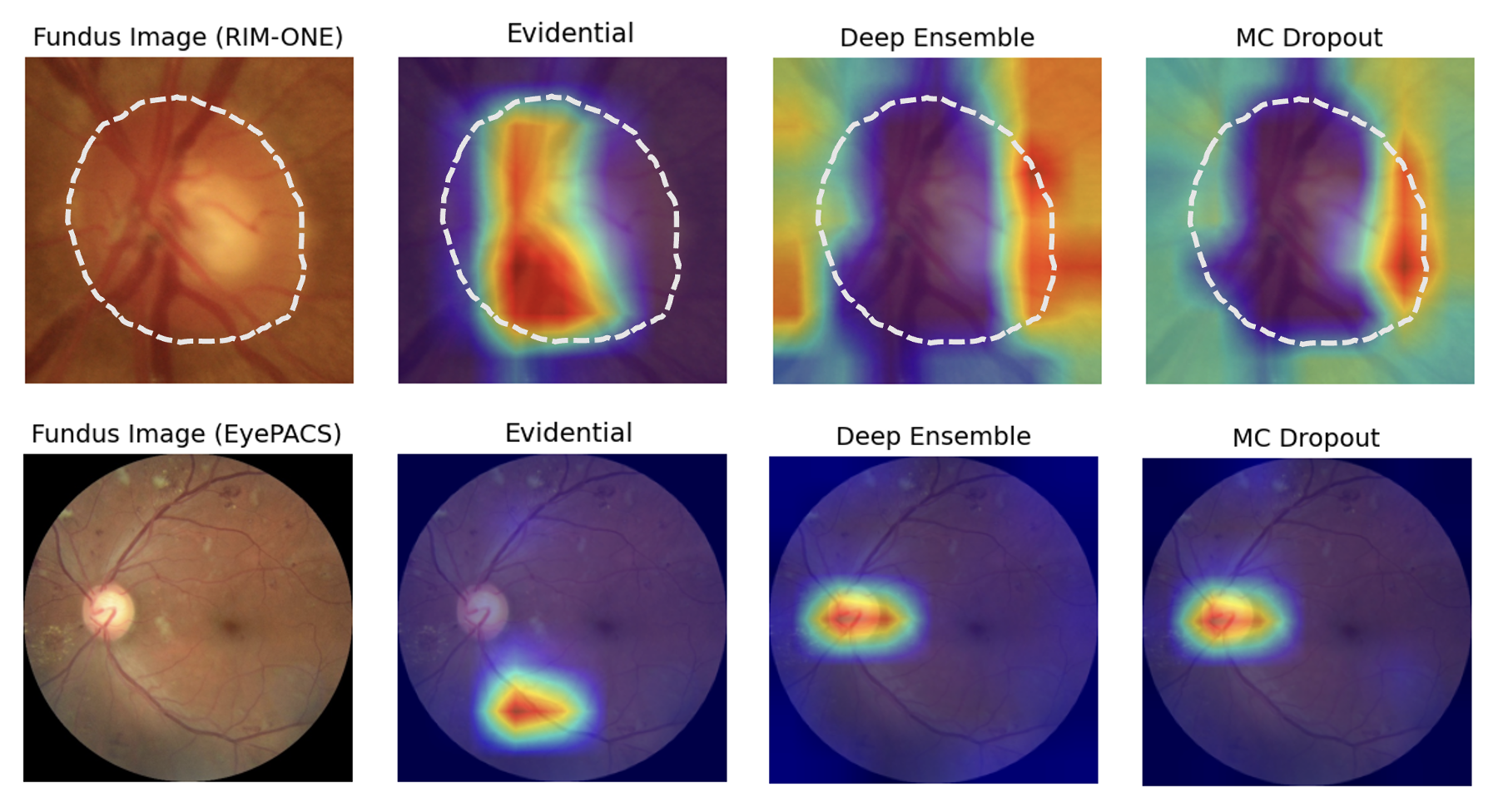}
    \caption{Grad-CAM visualizations for representative false negative fundus (leftmost) images of each dataset (\emph{top}: RIM-ONE-trained models, \emph{bottom}: EyePACS-trained models). In the top row, a dashed white contour delineates the optic disc. }
\label{fig:origa_fundus}
\end{figure}

In Fig.~\ref{fig:origa_fundus}, we show two representative samples of false negative classification. 
The heatmaps appeared to show that for the EDL model (for illustration, we chose the one trained with $\mathcal{L}_c$), the activation foci was visibly misplaced from the critical optic disc region yet much more localized compared to those of Deep Ensemble and MC Dropout. For the RIM-ONE image in Fig.~\ref{fig:origa_fundus} where the fundus image was already cropped around the optic disc region (conventionally an essential area to examine for glaucoma), the misplaced attention was still contained within the optic disc and overlapping with the neuro-retinal rim in contrast to the much more diffuse model attention generated by the other two models. For the EyePACS image in Fig.~\ref{fig:origa_fundus}, the misplaced attention of the EDL model was relatively more localized and the heatmap's misplaced attention was consistent with the spatial deviation from the optic disc neighborhood. This feature can be used by the clinician as one mechanism to help flag potential false negatives by leveraging on our understanding of glaucoma symptoms being mainly localized within the optic disc. In contrast, we noted that the attention areas of all other models other than EDL ones generally overlapped with the optic disc region. We found this distinguishing feature of misplaced attention mirroring false negative cases to characterize EDL models, regardless of whether they were trained with $\mathcal{L}_c$ or $\mathcal{L}_{kl}$. 

In this work, we have only used MobileNet as the backbone architecture. It would be important to examine other models, such as vision transformers, to study how universal the effectiveness of conditional entropy is. In the clinical operation setting, an uncertainty threshold associated with the minimal levels of sensitivity and precision can potentially be used to characterize model reliability. Thus, various types of accuracy-uncertainty threshold plots would be useful indicators to study prior to clinical implementation\footnote{We thank an anonymous reviewer for this suggestion.}. Another future direction would be to study if it is possible to characterize the nature of conditional entropy more precisely, for example whether it is representative of aleatoric or epistemic uncertainty \cite{depeweg,kendall2017}. In \cite{Gales}, a class of neural networks called \emph{Prior Networks} were proposed which also leveraged the conditional entropy we derived here. We note that these models were trained using loss and regularization terms significantly different from ours, and that although mentioned in their work (see Appendix C), the conditional entropy was not examined directly in their experiments. 
It would be interesting to study the other notions of entropy presented in \cite{Gales} in our context.

\section{Conclusion}
\label{sec:conclusion}

Our study of uncertainty-aware models for glaucoma classification has shown that, when built upon a pre-trained MobileNet backbone, EDL models incorporating our conditional entropy function (Eq.~\eqref{sm}), as implemented through the corresponding regularizer (Eq.~\eqref{L_m}), produced uncertainty estimates with substantially lower calibration errors than those obtained using Deep Ensembles, post-hoc temperature scaling, MC Dropout, and traditional EDL models based on Dempster uncertainty. Towards elucidating the interpretability of our models, we have also examined their Grad-CAM heatmaps. Distributions of Grad-CAM focus indices were relatively more discriminatory between correctly and wrongly classified fundus images
for EDL models compared to the other UQ methods. In particular, false negatives were associated with localized Grad-CAM activations that were displaced from the optic disc, consistent with its established physiological importance in glaucoma assessment. Overall, our findings suggest that our proposed EDL model based on conditional entropy provides not only enhanced uncertainty calibration, but also admits a more physiologically consistent interpretability, making it a potentially effective model for uncertainty-aware glaucoma detection from fundus images.


\begin{credits}
\subsubsection{\ackname} 
We express our gratitude to ophthalmologist Dr Claudine Pang of Asia Retina Eye Centre for various medical physiology-related explanations. We also thank the anonymous reviewers of TAAI 2026 for various feedback.

\subsubsection{\discintname}
The author has no competing interests to declare that are
relevant to the content of this article. 
\end{credits}
%
%
%

%
\appendix

\section{On the asymptotic behavior of conditional entropy $S_c$}
In this Appendix, we show that in the asymptotic region $\alpha, \beta \rightarrow \infty$, the conditional entropy function can be mapped to the standard binary entropy function. 
We begin by noting that the domain $((\alpha,\beta)\in[1,\infty)\times[1,\infty))$ can be decomposed into rays of fixed ratio described by
$
\beta=r\alpha,\, r>0.
$
We use this decomposition to examine the asymptotic behavior of the conditional entropy function $S_c$ along different directions in the $(\alpha,\beta)$-plane. 
Along each line with a fixed $r$,
\be
f_r (\alpha ) \equiv S_c(\alpha, r\alpha) = \psi \left(
(r+1)\alpha + 1
\right) - \frac{1}{1+r} \left(
\psi(\alpha + 1) + r\psi(r\alpha + 1)
\right).
\ee
Using $
\lim_{z\rightarrow \infty} \psi(z)
 \sim \log(z) - \frac{1}{2z}$,
we obtain 
\be
\label{mapping}
\lim_{\alpha \rightarrow \infty}
f_r(\alpha) \sim
\log \left( (r+1)\alpha + 1
\right) - \frac{1}{2((r+1)\alpha + 1)} 
\sim \overline{p} \log \overline{p} + (1-\overline{p}) \log (1-\overline{p} ),
\ee
where we have defined $\overline{p} = \frac{\alpha}{\alpha + \beta}$. This is the binary entropy function evaluated at $p = \overline{p}$. Another way of understanding this mapping is to note that the asymptotic region $\alpha, \beta \rightarrow \infty$ is where the Beta prior distributions reduce to singular $\delta$-functions centered at various values of $\overline{p}$. To see this, we introduce a scaling factor $s$ to write $\alpha = \tilde{\alpha}s, \beta = \tilde{\beta}s$, and invoke the following known relation~\cite{formula} between Beta and normal distributions which reads
\be
\label{beta_normal}
\lim_{s \rightarrow \infty}
\text{Beta} (\tilde{\alpha}s, \tilde{\beta}s ) \rightarrow 
\mathcal{N} \left(
\frac{\tilde{\alpha}}{\tilde{\alpha} + \tilde{\beta}}
, \frac{\tilde{\alpha}\tilde{\beta}}{(\tilde{\alpha} + \tilde{\beta})^3 s}
\right).
\ee
At any fixed $r$ (and thus $\overline{p}$), 
$$
\lim_{\substack{\alpha=r\beta \\ \alpha \to\infty}} \text{Beta}
(\alpha, \beta ) \rightarrow \lim_{\alpha \rightarrow \infty}
\mathcal{N} \left(
\overline{p}
, (1-\overline{p})\frac{1}{\alpha}
\right).$$ In this limit, there is no stochasticity in $p$, and thus the averaged entropy is simply the entropy evaluated at $\overline{p}$, consistent with what we have shown in eqn.~\eqref{mapping}.

\bibliographystyle{splncs04}

\bibliography{glau_ref}

@misc{Gales,
      title={Predictive Uncertainty Estimation via Prior Networks}, 
      author={Andrey Malinin and Mark Gales},
      year={2018},
      eprint={1802.10501},
      archivePrefix={arXiv},
      primaryClass={stat.ML},
      url={https://arxiv.org/abs/1802.10501}, 
}

@misc{kendall2017,
      title={What Uncertainties Do We Need in Bayesian Deep Learning for Computer Vision?}, 
      author={Alex Kendall and Yarin Gal},
      year={2017},
      eprint={1703.04977},
      archivePrefix={arXiv},
      primaryClass={cs.CV},
      url={https://arxiv.org/abs/1703.04977}, 
}

@misc{depeweg,
      title={Decomposition of Uncertainty in Bayesian Deep Learning for Efficient and Risk-sensitive Learning}, 
      author={Stefan Depeweg and José Miguel Hernández-Lobato and Finale Doshi-Velez and Steffen Udluft},
      year={2018},
      eprint={1710.07283},
      archivePrefix={arXiv},
      primaryClass={stat.ML},
      url={https://arxiv.org/abs/1710.07283}, 
}

@InProceedings{temperature,
  title     = {On Calibration of Modern Neural Networks},
  author    = {Guo, Chuan and Pleiss, Geoff and Sun, Yu and Weinberger, Kilian Q.},
  booktitle = {Proceedings of the 34th International Conference on Machine Learning},
  pages     = {1321--1330},
  year      = {2017},
  editor    = {Precup, Doina and Teh, Yee Whye},
  volume    = {70},
  series    = {Proceedings of Machine Learning Research},
  publisher = {PMLR}
}

@book{thomas,
  title     = {Elements of Information Theory},
  author    = {Cover, Thomas M. and Thomas, Joy A.},
  year      = {2006},
  edition   = {2nd},
  publisher = {Wiley-Interscience},
  address   = {Hoboken, N.J.},
  isbn      = {978-0-471-24195-9}
}

@inproceedings{amini,
  title={Deep Evidential Regression},
  author={Amini, Alexander and Schwarting, Wilko and Soleimany, Ava and Rus, Daniela},
  booktitle={Advances in Neural Information Processing Systems},
  volume={33},
  pages={14927--14937},
  year={2020}
}

@article{acrima,
	author = {Diaz-Pinto, Andres and Morales, Sandra and Naranjo, Valery and K{\"o}hler, Thomas and Mossi, Jose M. and Navea, Amparo},
	journal = {BioMedical Engineering OnLine},
	number = {1},
	pages = {29},
	title = {CNNs for automatic glaucoma assessment using fundus images: an extensive validation},
	volume = {18},
	year = {2019}}

@book{formula,
  title     = {Handbook of Mathematical Functions with Formulas, Graphs, and Mathematical Tables},
  author    = {Abramowitz, Milton and Stegun, Irene A.},
  year      = {1965},
  publisher = {Dover Publications},
  address   = {New York},
  edition   = {Dover reprint},
  isbn      = {978-0-486-61272-0}
}

@misc{hassan,
  author = {Hassan, Taimur and Akram, Muhammad Usman and Nazir, Muhammad Noman},
  title  = {A Composite Retinal Fundus and OCT Dataset with Detailed Clinical Markings of Retinal Layers and Retinal Lesions to Grade Macular and Glaucomatous Disorders},
  year      = {2021},
  publisher = {Mendeley Data},
  version   = {V4},
  doi       = {10.17632/trghs22fpg.4}
}

@misc{jungo,
      title={Assessing Reliability and Challenges of Uncertainty Estimations for Medical Image Segmentation}, 
      author={Alain Jungo and Mauricio Reyes},
      year={2019},
      eprint={1907.03338},
      archivePrefix={arXiv},
      primaryClass={eess.IV},
      url={https://arxiv.org/abs/1907.03338}, 
}

@InProceedings{gal,
  title = 	 {Dropout as a Bayesian Approximation: Representing Model Uncertainty in Deep Learning},
  author = 	 {Gal, Yarin and Ghahramani, Zoubin},
  booktitle = 	 {Proceedings of The 33rd International Conference on Machine Learning},
  pages = 	 {1050-1059},
  year = 	 {2016},
  editor = 	 {Balcan, Maria Florina and Weinberger, Kilian Q.},
  volume = 	 {48},
  series = 	 {Proceedings of Machine Learning Research},
  address = 	 {New York, New York, USA},
  month = 	 {20--22 Jun},
  publisher =    {PMLR}}

@inproceedings{lak, 
author = {Lakshminarayanan, Balaji and Pritzel, Alexander and Blundell, Charles}, title = {Simple and scalable predictive uncertainty estimation using deep ensembles}, year = {2017}, isbn = {9781510860964}, publisher = {Curran Associates Inc.}, 
booktitle = {Proceedings of the 31st International Conference on Neural Information Processing Systems}, pages = {6405–6416}, numpages = {12}, location = {Long Beach, California, USA}, series = {NIPS'17} }

@INPROCEEDINGS {FUM,
author = { Sanchez, J. L. and Sigut, J. and Gonzalez-Hernandez, M. and Alayon, S. and Fumero, F. },
booktitle = { Proceedings of the 26th IEEE International Symposium on Computer-Based Medical Systems },
title = {{ RIM-ONE: An open retinal image database for optic nerve evaluation }},
year = {2011},
volume = {},
ISSN = {},
pages = {1-6},
doi = {10.1109/CBMS.2011.5999143},
url = {https://doi.ieeecomputersociety.org/10.1109/CBMS.2011.5999143},
publisher = {IEEE Computer Society},
address = {Los Alamitos, CA, USA},
month =Jun}

@article{E2,
  author  = {Steen, Jessica and Kiefer, Riley and Ardali, Mahsa and Abid, Muhammad and Amjadian, Ehsan},
  title   = {Standardized and Open-Access Glaucoma Dataset for Artificial Intelligence Applications},
  journal = {Investigative Ophthalmology \& Visual Science},
  year    = {2023},
  volume  = {64},
  number  = {8},
  pages   = {384}
}

@misc{uq_review,
      title={A Survey on Uncertainty Quantification Methods for Deep Learning}, 
      author={Wenchong He and Zhe Jiang and Tingsong Xiao and Zelin Xu and Yukun Li},
      year={2025},
      eprint={2302.13425},
      archivePrefix={arXiv},
      primaryClass={cs.LG},
      url={https://arxiv.org/abs/2302.13425}, 
}

@article{mD,
	author = {Leng, Xiangjie and Shi, Ruijie and Wu, Yanxia and Zhu, Shiyin and Cai, Xingcan and Lu, Xuejing and Liu, Ruobing},
	journal = {PLoS One},
	number = {4},
	pages = {e0284060},
	title = {Deep learning for detection of age-related macular degeneration: A systematic review and meta-analysis of diagnostic test accuracy studies.},
	volume = {18},
	year = {2023}}

@article{dR,
	author = {Akhtar, Samia and Aftab, Shabib and Ali, Oualid and Ahmad, Munir and Khan, Muhammad Adnan and Abbas, Sagheer and Ghazal, Taher M.},
	journal = {Scientific Reports},
	number = {1},
	pages = {3763},
	title = {A deep learning based model for diabetic retinopathy grading},
	volume = {15},
	year = {2025}}

@article{Dulani,
	author = {Dulani Meedeniya and Thisara Shyamalee and Gilbert Lim and Pratheepan Yogarajah},
	journal = {Informatics in Medicine Unlocked},
	pages = {101644},
	title = {Glaucoma identification with retinal fundus images using deep learning: Systematic review},
	volume = {56},
	year = {2025}}

@article{Camara,
	author = {Camara, Jos{\'e} and Neto, Alexandre and Pires, Ivan Miguel and Villasana, Mar{\'\i}a Vanessa and Zdravevski, Eftim and Cunha, Ant{\'o}nio},
	journal = {J Imaging},
	month = {Jan},
	number = {2},
	title = {Literature Review on Artificial Intelligence Methods for Glaucoma Screening, Segmentation, and Classification.},
	volume = {8},
	year = {2022}}

@book{shields,
	address = {Philadelphia},
	author = {Allingham, R. Rand. and Shields, M. Bruce.},
	edition = {6th edition / R. Rand Allingham et al.},
	publisher = {Wolters Kluwer/Lippincott Williams and Wilkins Health},
	title = {Shields' textbook of glaucoma},
	year = {2011}}

@article{RIM_ONE, place={Ljubljana, Slovenia}, title={RIM-ONE DL: A Unified Retinal Image Database for Assessing Glaucoma Using Deep Learning}, volume={39}, url={https://www.ias-iss.org/ojs/IAS/article/view/2346}, DOI={10.5566/ias.2346}, number={3}, journal={Image Analysis and Stereology}, author={Fumero Batista, Francisco José and Diaz-Aleman, Tinguaro and Sigut, Jose and Alayon, Silvia and Arnay, Rafael and Angel-Pereira, Denisse}, year={2020}, month={Nov.}, pages={161–167} }

@inproceedings{GradCAM,
  title={Grad-CAM: Visual Explanations from Deep Networks via Gradient-based Localization},
  author={Selvaraju, Ramprasaath R. and Cogswell, Michael and Das, Abhishek and Vedantam, Ramakrishna and Parikh, Devi and Batra, Dhruv},
  booktitle={Proceedings of the IEEE International Conference on Computer Vision (ICCV)},
  pages={618--626},
  year={2017}
}

@inproceedings{PyTorch,
  author    = {Adam Paszke and Sam Gross and Francisco Massa and Adam Lerer and
               James Bradbury and Gregory Chanan and Trevor Killeen and
               Zeming Lin and Natalia Gimelshein and Luca Antiga and
               Alban Desmaison and Andreas Kopf and Edward Yang and
               Zachary DeVito and Martin Raison and Alykhan Tejani and
               Sasank Chilamkurthy and Benoit Steiner and Lu Fang and
               Junjie Bai and Soumith Chintala},
  title     = {PyTorch: An Imperative Style, High-Performance Deep Learning Library},
  booktitle = {Advances in Neural Information Processing Systems (NeurIPS)},
  volume    = {32},
  year      = {2019}
}

@inproceedings{Heming,
  author    = {Kaiming He and Xiangyu Zhang and Shaoqing Ren and Jian Sun},
  title     = {Delving Deep into Rectifiers: Surpassing Human-Level Performance on ImageNet Classification},
  booktitle = {Proceedings of the IEEE International Conference on Computer Vision (ICCV)},
  year      = {2015},
  pages     = {1026--1034},
  doi       = {10.1109/ICCV.2015.123}
}

@article{MobileNetV3,
  author  = {Andrew Howard and Mark Sandler and Grace Chu and Liang-Chieh Chen and
             Bo Chen and Mingxing Tan and Weijun Wang and Yukun Zhu and
             Ruoming Pang and Vijay Vasudevan and Quoc V. Le and Hartwig Adam},
  title   = {Searching for MobileNetV3},
  journal = {Proceedings of the IEEE/CVF International Conference on Computer Vision (ICCV)},
  year    = {2019},
  pages   = {1314--1324}
}

@ARTICLE{DS,
  author={A. P. Dempster},
  journal={Journal of the Royal Statistical Society }, 
  title={A Generalization of Bayesian Inference}, 
  year={1968},
  volume={B},
  number={10},
  pages={205-247}}

@book{shafer,
  title={A Mathematical Theory of Evidence},
  author={Shafer, Glenn},
  year={1976},
  publisher={Princeton University Press},
  address={Princeton, NJ}
}

@article{Anderson,
title = {Absence of Diffusion in Certain Random Lattices},author = {Anderson, P. W.},journal = {Physical Review},volume = {109},number = {5},pages = {1492--1505},year = {1958},month = {March},doi = {10.1103/PhysRev.109.1492}}

@misc{sensoy,
      title={Evidential Deep Learning to Quantify Classification Uncertainty}, 
      author={Murat Sensoy and Lance Kaplan and Melih Kandemir},
      year={2018},
      eprint={1806.01768},
      archivePrefix={arXiv},
      primaryClass={cs.LG},
      url={https://arxiv.org/abs/1806.01768}, 
}

@misc{eyepacs2,
	title={Glaucoma Dataset: EyePACS-AIROGS-light-V2},
	url={https://www.kaggle.com/dsv/7802508},
	publisher={Kaggle},
	author={Riley Kiefer},
	year={2024}
}

@article{airogs,
  title={AIROGS: Artificial Intelligence for RObust Glaucoma Screening Challenge},
  author={de Vente, Coen and Vermeer, Koenraad A. and Jaccard, Nicolas and Wang, He and Sun, Hongyi and Khader, Firas and others},
  journal={arXiv preprint arXiv:2302.01738},
  year={2023}
}

\end{document}